\documentclass[lettersize,journal]{IEEEtran}
\usepackage{amsmath,amsfonts}
\usepackage{algpseudocode}
\usepackage{algorithm}
\usepackage{array}
\usepackage[caption=false,font=normalsize,labelfont=sf,textfont=sf]{subfig}
\usepackage{textcomp}
\usepackage{stfloats}
\usepackage{url}
\usepackage{verbatim}
\usepackage{graphicx}
\usepackage{cite}
\usepackage{booktabs}
\usepackage{siunitx}
\usepackage{multirow}
\usepackage{makecell}
\usepackage{xcolor} 
\usepackage{threeparttable}
\usepackage{ulem}
\begin{document}

\title{Multi-Representation Geometric Hierarchy Fusion: An Implicit-Submap Driven Framework for Resilient 3D Place Recognition}

\author{Xiaohui Jiang, Haijiang Zhu, Chade Li, Ning An*
\thanks{This work was supported in part by the National Natural Science Foundation of China (Grant No.62202468 and No. 92367111) and the Key Science and Technology Innovation Project of CCTEG (No. 2024-TD-ZD016-01, 2024-TD-MS017). (corresponding author: Ning An)}
\thanks{Xiaohui Jiang is with the College of Information and Technology, Beijing University of Chemical Technology, Beijing, 100029,    P. R. China. and also with the School of Artificial Intelligence and Robotics, Hunan University, Changsha,410082, P. R. China. (e-mail: jiangxiaohui2001@gmail.com)}
\thanks{Haijiang Zhu is with the College of Information and Technology, Beijing University of Chemical Technology, Beijing 100029, P. R. China. (e-mail: zhuhj@mail.buct.edu.cn)}
\thanks{Chade Li, is with the ICT BG, Beijing Research Institute, Huawei Technologies Co., Ltd., Beijing, 100095, P. R. China (e-mail: lichade2@huawei.com, chad\_lee\_richard@outlook.com)}
\thanks{Ning An is with the Research Institute of Mine Artificial Intelligence, China Coal Research Institute, Beijing, 100013, P.R. China. and also with the State Key Laboratory of Intelligent Coal Mining and Strata Control, Beijing 100013, P. R. China. (e-mail:ning.an@ccteg-bigdata.com)}
}

\markboth{Journal of \LaTeX\ Class Files,~Vol.~14, No.~8, August~2021}%
{Shell \MakeLowercase{\textit{et al.}}: A Sample Article Using IEEEtran.cls for IEEE Journals}


\maketitle

\begin{abstract}
LiDAR-based place recognition is critical for long-term autonomous driving without GPS. Existing handcrafted feature methods face dual limitations. First, descriptor instability occurs due to inconsistent point cloud density from motion and environmental changes during repeated traversals. Second, representation fragility arises from reliance on single-level geometric abstractions in complex scenes. To overcome these, we propose a novel framework for 3D place recognition. We introduce an implicit 3D representation using elastic neural points. This representation is designed to reduce the influence of input-density variations and to provide more regular geometric evidence for descriptor construction. From this, we derive occupancy grids and normal vectors. These enable the construction of fused descriptors that integrate complementary perspectives: macro-level spatial layouts from a bird's-eye view and micro-scale surface geometries from 3D clusters. Extensive evaluations on diverse datasets, including KITTI, KITTI-360, MulRan, and NCLT, demonstrate that the proposed method achieves competitive and robust performance compared with representative handcrafted and learning-based baselines. The results suggest that the proposed framework provides a favorable trade-off among recognition accuracy, runtime efficiency, and map memory footprint. It also shows improved robustness under density variations and viewpoint changes in the evaluated scenarios. The code will be released soon.
\end{abstract}


\section{Introduction}
\IEEEPARstart{P}{LACE} recognition is the ability to identify previously visited locations in robotics and autonomous driving systems. It serves as a cornerstone for achieving long-term autonomy in transportation systems \cite{ref1}. This capability is crucial for autonomous mobile robots to achieve precise and robust positioning in unknown environments \cite{ref2}. Place recognition also has a wide range of applications. For example, it is used in loop closure detection for Simultaneous Localization and Mapping (SLAM) \cite{ref4,ref5}, and autonomous robot navigation \cite{ref8}. 
Light Detection and Ranging (LiDAR) can obtain accurate measurement results even under various lighting conditions, such as in darkness or when disturbed by strong light. As a result, LiDAR plays a crucial role in unmanned systems. Meanwhile, with the rapid development of autonomous and robotic driving systems, there is an increasing demand for LiDAR-based place recognition \cite{ref10,ref11}. 

\begin{figure}[!t]
    \centering
    \includegraphics[scale=0.3]{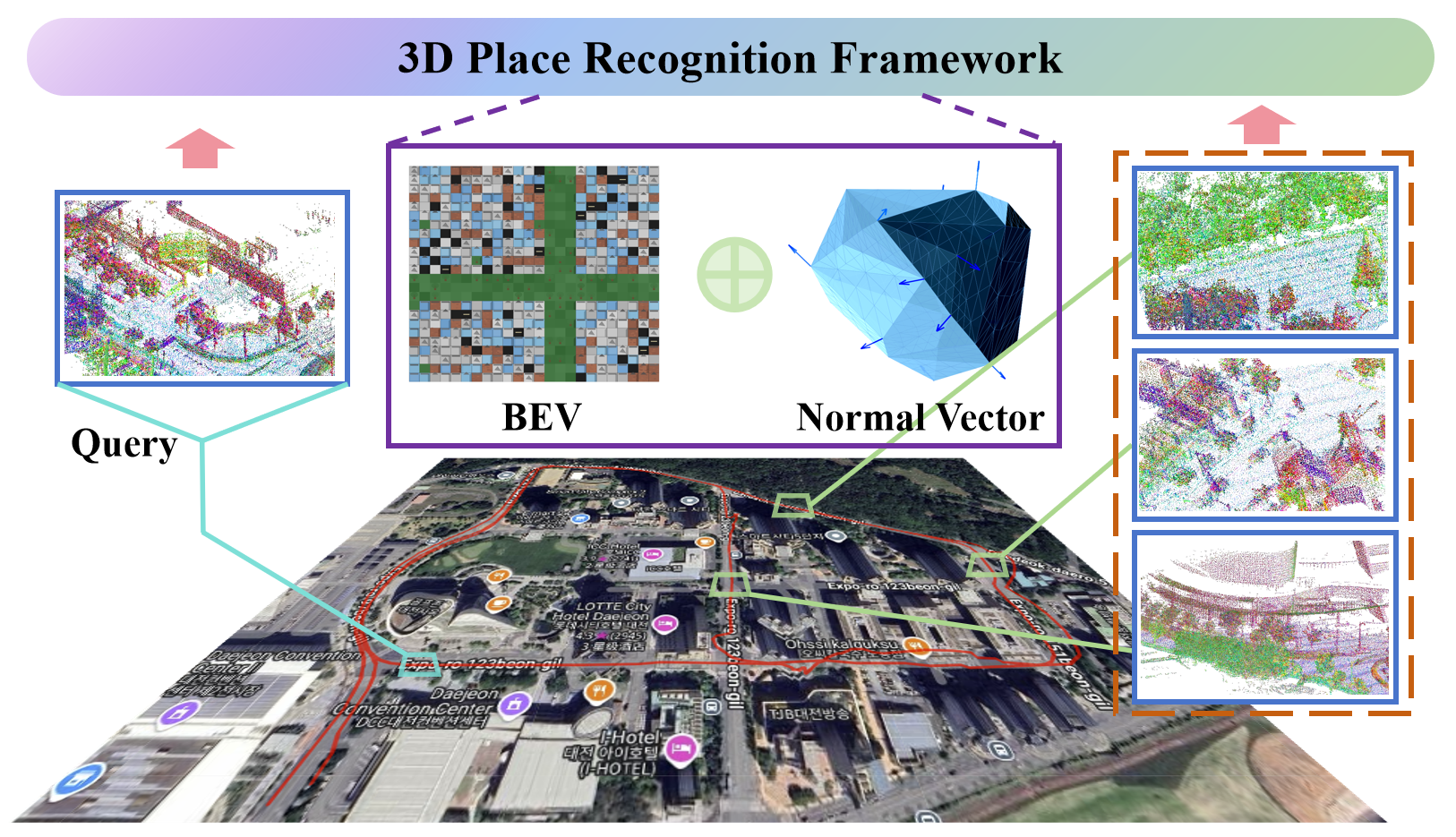}
    \caption{Our pipeline first converts multi-frame submaps into an elastic neural-point implicit representation, from which we derive occupancy evidence and consistent normals to build macro–micro descriptors. The colored dots in the graph represent these neural representation points. Subsequently, utilizing the advantages of implicit representations, we fuse information from two complementary geometric representations/cues, namely BEV occupancy-layout features and surface-normal statistics of 3D clusters, to generate descriptors for each submap.}
    \label{fig_8}
\end{figure}
In recent years, methods for place recognition based on LiDAR point clouds have developed rapidly. However, these methods still face several key challenges in practical applications: 
\begin{itemize}
\item[$\bullet$] In urban environments, LiDAR point clouds from the same location often differ in density and coverage. These inconsistencies are due to variations in the trajectories and speeds of vehicles or robots. The algorithm, therefore, must be robust to such variability. It needs to maintain rotational and translational invariance, as well as invariance to point cloud density \cite {ref13,ref14}. 
\item[$\bullet$] Dynamic objects in the scene, such as cars, pedestrians, and cyclists, constantly change the 3D structure of the environment. Their presence is unavoidable. This poses a challenge to the robustness of LiDAR point cloud-based place recognition \cite{ref15}.
\end{itemize}

Most current hand-crafted feature extraction-based 3D place recognition methods follow two main strategies.
One strategy extracts features (such as local features  \cite{ref17,ref18} or global features \cite{ref19,ref20}) directly from the raw 3D point cloud. 
The alternative technique first transforms the point cloud into a structured format (for example, Bird’s Eye View (BEV) \cite{ref21} or sparse voxel grids \cite{ref24}) before extracting features.

However, existing hand-crafted 3D feature extraction methods encounter a fundamental limitation: their dependence on narrow, single-representation geometric scene descriptions. These methods typically process either unstructured raw point clouds or a particular structured intermediate representation (e.g., BEV, Polar coordinates) derived from the point clouds. Critically, this reliance on any single representation fundamentally undermines feature robustness and coverage. Raw point clouds, although preserving exact geometric details, exhibit non-uniform spatial distributions (density variations) and lack intrinsic rotational and translational invariance, which hampers consistent feature extraction. In contrast, structured representations (e.g., BEV) enable concise global perspectives suitable for feature extraction, but they also forfeit detailed local geometry and vertical context, and perpetuate the point cloud's spatial irregularities. These inherited irregularities intensify the challenge of obtaining stable and discriminative features. Thus, features extracted solely from one representation remain insufficient for thorough and reliable 3D scene characterization, constraining downstream 3D place recognition performance.

Prior submap-based place recognition and hierarchical fusion methods have shown the benefit of aggregating multiple frames and combining complementary cues. However, most of them still build descriptors on explicit submap representations (raw points/voxel grids), where (i) surface normals estimated are sensitive to non-uniform sampling and density variations, and (ii) occupancy/BEV projections inherit discretization artifacts \cite{ref15,ref18}. As a result, fusion alone does not fully address descriptor instability under repeated traversals with different sampling patterns. Our work targets this missing link by constructing an implicit submap that yields stable occupancy evidence and consistent normals for descriptor formation.

To address these issues, we have moved away from the previous explicit 3D point cloud input and, instead, now use an elastic neural points-based implicit representation. This representation is introduced for three key reasons that are outlined below.
\begin{itemize}
\item[$\bullet$] The elastic neural points-based implicit representation is compact, flexible, and continuous. It possesses rotational and translational invariance, as well as spatial density invariance.  
\item[$\bullet$] This representation can efficiently provide high-quality, globally consistent surface normals, offering the opportunity to derive more discriminative place recognition descriptors.
\item[$\bullet$] This representation enables more flexible online removal of dynamic objects, leading to more robust place recognition.
\end{itemize} 

We propose a multi-representation macro–micro descriptor construction and fusion scheme built on an implicit submap: macro BEV saliency guides the computation budget and sampling of micro cluster normal statistics, rather than performing a uniform extraction or a simple feature concatenation.
As illustrated in Fig.~\ref{fig_8}, our method uses a point-based implicit scene representation, combining Signed Distance Function (SDF) and stability prediction to filter dynamic objects online. We synergistically fuse macro-level Bird's-Eye View (BEV) descriptors, which capture global structural context, with micro-scale normal-based descriptors that encode distinctive local geometry. Specifically, features from BEV descriptors guide the identification and interpretation of salient local features, while the detailed information from normal-based descriptors refines and strengthens the BEV-based scene understanding. This bidirectional guidance enables each descriptor type to enhance the other's contribution, resulting in a robust and comprehensive scene descriptor that supports enhanced place recognition.

In summary, the contributions of this article are as follows:
\begin{itemize}
\item[$\bullet$]
We introduce an implicit-submap-to-geometry pipeline using elastic neural points, from which we derive density-invariant occupancy evidence and globally consistent normals for place recognition.
\item[$\bullet$] 
We propose a saliency-scheduled macro–micro geometric-cue fusion: BEV keypoints allocate micro-level cluster normal-statistics evaluations, producing a fused descriptor that is more robust than simple concatenation.
\item[$\bullet$]We present a descriptor construction scheme that does not require place-recognition supervision. The implicit submap is optimized online using self-supervised SDF regression, while the final place-recognition descriptor is built without PR labels. Experiments show that the proposed method achieves competitive performance with favorable storage efficiency.
\end{itemize} 

\section{Related Works}
In this section, we provide a comprehensive review of implicit map representation, as well as 3D point cloud place recognition methods based on both handcrafted and deep learning approaches.  

\subsection{Implicit Neural Map Representation}
Explicitly represented maps (e.g., point clouds\cite{ref27}, meshes\cite{ref28}, voxel/occupancy grids\cite{ref30}) have been widely used in unmanned systems. Recently, implicit neural representations have demonstrated effectiveness in modelling radiation\cite{ref31} and geometric fields\cite{ref32,ref33}, enabling applications such as 3D reconstruction and novel view synthesis. Methods such as NeRF\cite{ref31}, DeepSDF\cite{ref33}, and occupancy networks\cite{ref32} typically employ a single MLP to represent entire scenes. While effective, this approach faces efficiency bottlenecks (scalability, computation, and memory) in complex scenes. To address this, recent work has adopted hybrid representations that jointly optimize explicitly stored local latent features with shallow multilayer perceptrons (MLPs) \cite{ref34,ref35}.

Advances in efficient training now enable the scaling of implicit neural representations to larger scenes, accelerating their adoption in mapping and SLAM. For RGB-D SLAM, two main approaches exist for simultaneous scene modeling and camera pose tracking: one uses a single MLP to represent the entire scene as a global implicit function \cite{ref38}, while the other uses grid-based local latent features with multiple shallow MLPs, enabling local modelling and improved scalability\cite{ref41}.

In addition to RGB-D approaches, LiDAR-based systems have also advanced, with methods like IR-MCL\cite{ref42} and LocNDF\cite{ref43} constructing implicit neural distance maps for robot localization. 
Building upon these developments, SHINE-mapping \cite{ref44} extends implicit neural representations to large-scale outdoor LiDAR data using octree-based sparse voxel features for efficient storage. NeRF-LOAM\cite{ref46} implements an online-optimizable octree feature grid, enabling efficient LiDAR odometry and mapping. PIN-SLAM\cite{ref47} employs elastic neural points with hash-table indexing, significantly improving real-time performance and storage efficiency over previous neural mapping techniques. These LiDAR-based works demonstrate complementary advantages of implicit neural representations, including spatial continuity and rich feature encoding. Inspired by these distinctions, our method employs a point-based implicit neural representation. From this continuous representation, we directly derive scene surface normals and occupancy information, forming the essential basis for subsequent place recognition.

\subsection{Handcrafted Methods}
Early studies primarily described point clouds by extracting their geometric local features, including point normal \cite {ref49}, angle \cite {ref50}, among others. These methods have achieved success in point cloud registration and shape recognition. As research progressed, researchers began to encode the entire point cloud into a global feature descriptor. For example,  M2DP \cite{ref54} projects the point cloud onto multiple two-dimensional planes, calculates the point density on each plane, and forms a signature vector. This method avoids calculating point normals, thereby improving computational efficiency; however, some information may be lost. In contrast, Scan Context \cite {ref55,ref56} divides the horizontal space into multiple annular regions and fan-shaped regions, forming a two-dimensional height matrix. By performing nearest neighbor search through the annular key and the similarity score, good results have been achieved. To further enhance performance, researchers have begun to combine other types of features, such as intensity\cite{ref57} and spatial binary patterns\cite{ref58}, among others. For instance, LiDAR Iris\cite{ref59} converts point cloud data into "iris images", generates binary signatures through the LoG-Gabor filter and threshold operations, and then calculates the similarity through the Hamming distance. In another direction, STD\cite{ref18} proposes a stable triangle descriptor based on the invariance of the triangle shape. By extracting key points and encoding them into triangle descriptors, it enables efficient position matching and geometric verification. Additionally, BoW3D\cite{ref82} employs an innovative bag-of-words model to process 3D LiDAR features. Finally, RING++\cite{ref83} constructs a rotation and translation invariant representation with strict mathematical guarantees by Radon transform, Fourier transform, and cross correlation, and solves the global localization problem on sparse maps.

Most handcrafted feature methods utilize a single geometric representation, such as raw point clouds, voxel grids, or BEV maps, resulting in information loss as cues from other sources are overlooked. While simple, these methods struggle with major rigid transformations or complex scenes. To address this, our approach uses an implicit neural point representation that combines 3D cluster normals and BEV features. By hierarchically fusing these complementary geometric representations, we build a more robust and comprehensive representation for challenging scenarios.

\subsection{Learning-based Methods}
PointNetVLAD\cite{ref17} is a deep learning-based method that combines two crucial components: PointNet\cite{ref60} and NetVLAD\cite{ref61}. PointNet extracts features from point clouds. NetVLAD aggregates these local features into a 512-dimensional global descriptor. This method lays the foundation for subsequent approaches. 
Building on PointNetVLAD, SOE-Net~\cite{ref62} added attention-based feature encoding. In visual place recognition, hybrid CNN-transformer features with semantic NetVLAD aggregation have also shown the value of combining local detail and global context~\cite{ref63}. 
Locus\cite{ref11} enhances the descriptors by introducing temporal information. This addition improves robustness against viewpoint changes. HiTPR \cite{ref65}  further models local-global dependencies in voxelized point clouds, and Point Tree Transformer explores hierarchical attention for efficient point-cloud registration \cite{ref66}. 
Kim et al.\cite{ref67} proposed a classifier based on a Convolutional Neural Network (CNN) for long-term place recognition. This approach can handle significant scene changes over an extended period. Semantic Scan Context \cite {ref68} addresses the translation problem by introducing semantic information. 
LPD-Net\cite{ref69} uses graph-based feature aggregation; training-free random-network co-ensembles provide an alternative for robust registration under distribution shifts \cite{ref70}. 
DAH-Net\cite{ref71} proposes a density-driven adaptive hybrid network using density changes of point clouds. It combines dynamic local feature aggregation with a contrastively enhanced linear attention module. With these, it attempts to address the challenges caused by density variations in point clouds from large-scale scenes. LCDNet \cite {ref84} introduces a novel, unbalanced, optimal transport-based differentiable module. This module enables joint loop closure detection and 6-DoF pose estimation, allowing robust registration under arbitrary initial rotations and excelling at reverse loops. BEVPlace++\cite{ref85} introduces a Rotation Equivariant Module (REM), enabling weakly supervised, real-time LiDAR global localisation. It extracts rotation-equivariant features from Bird's Eye View images, achieving robust 3-DoF pose estimation without precise ground truth. These learning-based methods often require relatively long training times and substantial computational resources. Moreover, they perform poorly when facing scenarios that fall outside their training samples.  

\section{METHOD}
\begin{figure*}[!t]
\centering
\includegraphics[width=0.78\textwidth]{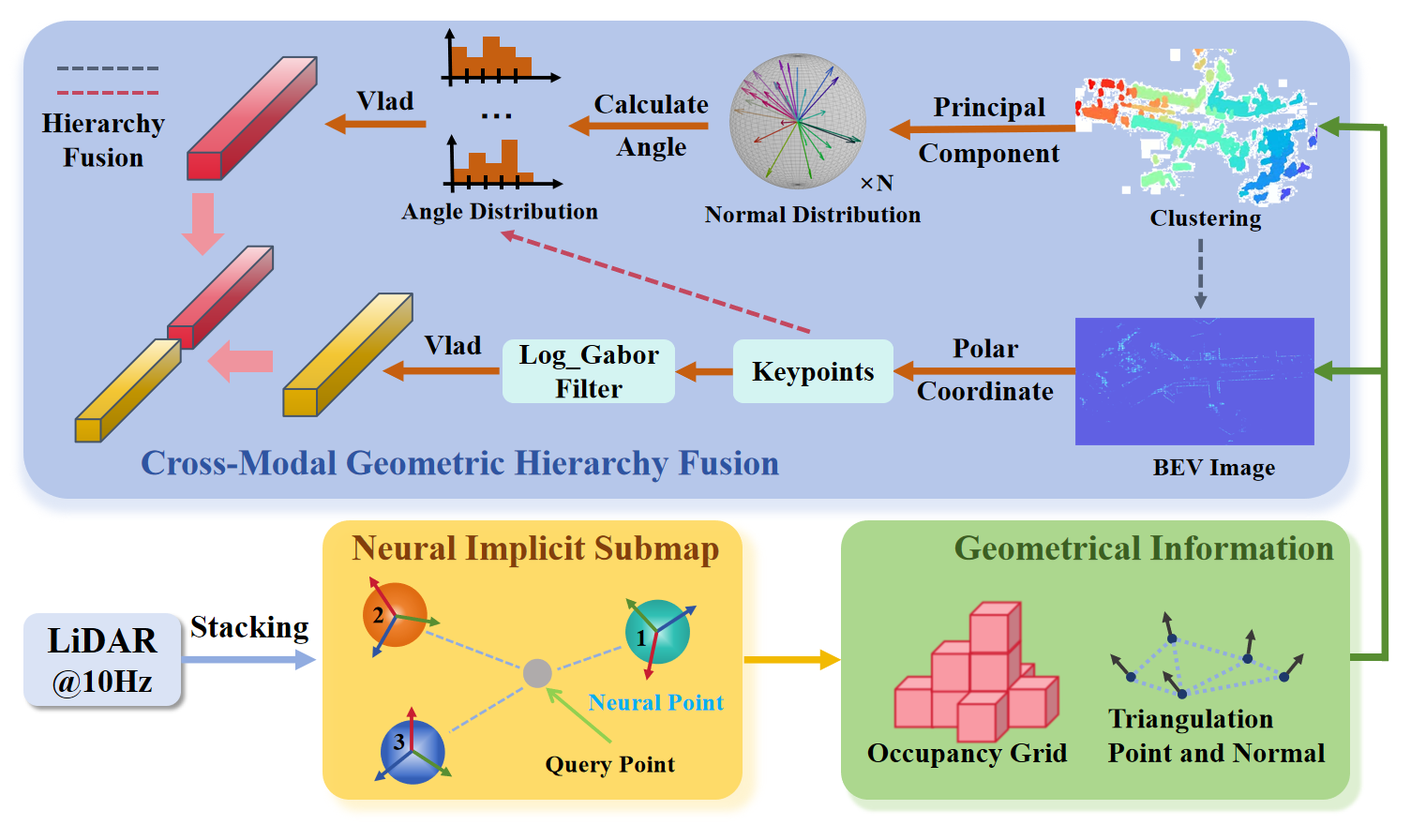}
\caption{This is the pipeline for our 3D Place Recognition. We feed a specific number of LiDAR frames to form a submap, which we then transform into an implicit neural point representation. From this, we derive high-quality occupancy grid information, triangular points, and normal vectors. BEV and primary 3D clusters are obtained next. We use a log-Gabor filter to generate descriptors for the BEV, while the angular difference distribution of normal vectors yields geometric descriptors for the primary 3D clusters. Finally, we combine these two descriptor types to form a single one for 3D place recognition. In the figure, yellow cuboids show macro descriptors from the bird's-eye view, and blue cuboids show micro descriptors from the main 3D clusters. The two descriptors are hierarchically fused to integrate their advantages.}
\label{fig_1}
\end{figure*}
This section introduces the proposed method for LiDAR-based place recognition. Section \uppercase\expandafter{\romannumeral3}-A provides an overview of the proposed framework. Section \uppercase\expandafter{\romannumeral3}-B presents the elastic neural implicit submap representation. Sections \uppercase\expandafter{\romannumeral3}-C and \uppercase\expandafter{\romannumeral3}-D describe the construction of the normal-vector descriptor and the BEV descriptor, respectively. Section \uppercase\expandafter{\romannumeral3}-E introduces the proposed multi-representation geometric hierarchy fusion strategy. 

\subsection{Overview}
The main process of our proposed method is illustrated in Fig.~\ref{fig_1}. To better handle the random noise caused by variations in the density of point clouds (which refers to fluctuations in how densely the 3D data points, collected by sensors, are distributed in space) and by dynamic objects (objects in the environment that move or change over time), and to obtain more discriminative descriptors (distinctive features used to tell apart different parts of a scene), we use elastic neural implicit 3D points. This representation models the environment as a continuous field using a neural network, allowing for the flexible positioning of points to more effectively describe the scene. 

Because LiDAR (a technology that measures distances using laser light to produce detailed 3D maps) point clouds acquired in different scanning modes need to be processed, our method takes as input several frames of point clouds $\mathbf{P_{1}},\dots,\mathbf{P_{n}}$, where each frame is a collection of 3D data points. The number $n$ depends on the distance the robot has travelled, denoted by $\tau_{n}$, which represents the path length corresponding to $n$ frames.

Next, for the input sequence of point cloud frames, we utilise a point-based implicit representation to encode the scene's geometry. Starting from the first frame, the system generates samples (selected points) and labels (information or categories assigned to samples) to optimise the neural implicit map, which results in a submap (a smaller localised map) representation. We then extract from this implicit submap: the scene's occupancy grid (a division of space into a grid where each cell indicates occupancy), sample points (vertices of a generated mesh approximating surfaces), and surface normals (vectors perpendicular to the surface at each mesh vertex). 

Then, we use Euclidean clustering (a technique that groups points based on how close they are to each other in 3D space) to quickly extract the main 3D clusters of the scene. These clusters guide the projection of the occupancy grid and the creation of the final bird’s-eye-view (BEV) representation—a top-down 2D view of the 3D space. We detect keypoints (distinct and important points in the data) on the BEV map, and calculate their spatial statistics (measures that summarise their distribution). These statistics provide initial information used to guide the extraction of feature descriptors (numeric vectors that describe local geometric properties) in the main 3D clusters, thus generating a dense set of micro-scale surface signatures $\mathbf{F^{G}} \in \mathbb{R}^{a}$. Next, we aggregate the BEV-derived features using unsupervised clustering (a method to find groups in unlabeled data), to produce a compact BEV descriptor $\mathbf{F^{B}} \in \mathbb{R}^{b}$ that is highly distinctive. Finally, both descriptors are combined into a single fused feature representation $\mathbf{F^{fuse}} \in \mathbb{R}^{a+b}$.

\subsection{Implicit Submap Representation With Elastic Neural Points}
We employ a memory-efficient implicit neural representation to model the geometry of each submap as a continuous signed distance field (SDF). Instead of explicitly storing dense geometric primitives, the submap incrementally accumulates local scene geometry through a set of elastic neural points, which serve as adaptive anchors for the implicit SDF representation. This design naturally supports incremental updates, local rotational and translational invariance, and robustness to variations in point cloud density. Moreover, by continuously re-parameterizing neural point attributes during submap accumulation, the representation is capable of suppressing interference from dynamic objects and maintaining long-term geometric consistency.

Here, the term elastic indicates that the neural points are not static entities: both their 3D spatial locations and associated latent features are dynamically updated as new observations are integrated into the submap.

\noindent1) \textit{Elastic Implicit Neural Point Cloud}: 
Within each submap, the accumulated geometry is represented by a collection of elastic implicit neural points, denoted as:
\begin{equation}
\label{eq1}
\mathcal{M}=\left \{\mathbf{ m_{i}} =\left ( \mathbf{p_{i}},\mathbf{q_{i}},\mathbf{f_{i}},t_{i}^{c},t_{i}^{u},\mu_{i} \right )  \mid i=1,\dots ,N\right \} 
\end{equation}
Each neural point $\mathbf{m_{i}}$ encodes local geometric information within the submap. Specifically, $\mathbf{p_{i}} \in \mathbb{R}^{3}$ and $\mathbf{q_{i}} \in \mathbb{R}^{4}$ represent the position and orientation of the neural point in the global coordinate frame. The optimizable latent feature $\mathbf{f_{i}} \in \mathbb{R}^{F}$ captures local surface characteristics and supports implicit SDF prediction. To facilitate submap maintenance and temporal consistency, each neural point additionally stores a creation timestamp $t_{i}^{c}$, a last update timestamp $t_{i}^{u}$, and a stability metric $\mu_{i}$, which reflects the persistence of the point across multiple observations. During submap accumulation, the set of neural points is dynamically maintained: neural points may be newly instantiated, updated, or replaced as additional sensor data is incorporated.

\noindent2) \textit{SDF Decoder}: 
To reconstruct a continuous signed distance field within the submap, we employ an implicit neural decoder that maps fused local encodings to SDF values. The decoder is incrementally adapted during the mapping process to reflect newly observed geometry. To ensure invariance of the SDF prediction to local translations and rotations of neural points, query positions are transformed from the global coordinate frame into the local frame of each neighboring neural point. Let $\mathbf{p_{s}} \in \mathbb{R}^{3}$ denote a query position in the global frame. For a neural point $\mathbf{m_{j}}$, the corresponding local coordinate $\mathbf{d_{j}} \in \mathbb{R}^{3}$ is computed as:

\begin{equation}
\label{eq2}
\mathbf{d_{j} } =\mathbf{q_{j}}( \mathbf{p_{s}}- \mathbf{p_{j}})\mathbf{q_{j}^{-1} },j=1,\dots ,K
\end{equation}
At each query position $\mathbf{p_{s}}$, the $K$ nearest neural points are retrieved from the neural point map, forming a local neighborhood $\mathcal{N}_{P}$. For each neural point $\mathbf{m{j}} \in \mathcal{N}_{P}$, a distance-based weight $\omega{j}$ is assigned as:
\begin{equation}
\label{eq3}
\omega _{j}=\frac{\left \| \mathbf{d_{j}}  \right \|^{-1}  }{\sum _{k\in \mathcal{N}_{P}  }\left \| \mathbf{d_{k}}  \right \|^{-1} }=\frac{\left \| \mathbf{p_{s}} -\mathbf{p_{j}}  \right \|^{-1}  }{\sum _{k\in \mathcal{N}_{P}  }\left \| \mathbf{p_{s}} -\mathbf{p_{k}}  \right \|^{-1}}
\end{equation}
Each neural point contributes both a feature encoding and a coordinate encoding. The coordinate encoding $\mathbf{g_{j}} \in \mathbb{R}^{C}$ is obtained via a positional encoding function $\gamma(\cdot)$ applied to $\mathbf{d_{j}}$, i.e., $\mathbf{g_{j}} = \gamma(\mathbf{d_{j}})$. The fused feature encoding $\mathbf{f} \in \mathbb{R}^{F}$ and coordinate encoding $\mathbf{g} \in \mathbb{R}^{C}$ for the query position are computed as weighted combinations over the local neighborhood:
\begin{equation}
\label{eq4}
\mathbf{f} =\sum _{j\in \mathcal{N}_{P}} \omega _{j} \mathbf{f_{j} },\mathbf{g} =\sum _{j\in \mathcal{N}_{P}} \omega _{j} \mathbf{g_{j}} 
\end{equation}

Finally, the implicit decoder $D_{\theta}$ predicts the signed distance value $s$ at the query position:
\begin{equation}
\label{eq5}
s=D_{\theta } (\mathbf{f},\mathbf{g})
\end{equation}
where $D_{\theta}$ is implemented as a shallow multilayer perceptron (MLP) with $M_{mlp}$ hidden layers and $N_{mlp}$ neurons per layer.

\noindent3) \textit{Neural Point Insertion and Maintenance}: 
During submap accumulation, upon acquisition of a new LiDAR point cloud, each point is evaluated for potential insertion into the neural point map. A point is instantiated as a new neural point if one of the following mutually exclusive conditions is satisfied: (1) The corresponding hash-table entry is empty, indicating that no neural point has previously been assigned to that spatial region; (2) The hash-table entry is occupied, but the Euclidean distance between the queried point $\mathbf{p}$ and the stored neural point exceeds a predefined threshold, indicating a spatial mismatch. In this case, the existing neural point is evicted and replaced by a newly initialized one. This strategy enables efficient spatial coverage while maintaining a bounded number of neural points within each submap.

\noindent4) \textit{Dynamic Point Filtering:} 
To further improve the robustness of the submap representation, we incorporate a dynamic point filtering mechanism based on the stability metric $\mu_{i}$ and SDF values. A point is classified as dynamic and excluded from the submap if its SDF value exceeds a dynamic distance threshold $\gamma_{d}$ and its stability exceeds a stability threshold $\gamma_{\mu}$. Specifically, a sampled point $\mathbf{p_{W}}$ in the world coordinate frame is considered dynamic if:
\begin{equation}
\label{eq41}
\mathbf{S}(\mathbf{p_{W}})> \gamma _{d}, \mathbf{H}(\mathbf{p_{W}})> \gamma _{\mu }
\end{equation}
where $\mathbf{S}(\mathbf{p_{W}})$ denotes the predicted SDF value, representing the distance to the nearest surface, and $\mathbf{H}(\mathbf{p_{W}})$ measures the temporal consistency of the point across observations. This filtering process operates at the submap level and effectively suppresses transient structures, thereby enhancing the accuracy and consistency of the implicit SDF representation.

\noindent5) \textit{Sample/Label Generation and Online Update:} Upon receiving a new LiDAR scan, we first transform the scan points into the current submap (world) frame using the estimated pose. We then build a lightweight training set for incremental SDF fitting by combining (i) on-surface samples taken from the measured returns with target SDF set to 0, and (ii) off-surface samples drawn along each LiDAR ray, where points in front of the return are labeled with positive distances to the surface and points slightly behind the return are labeled with negative distances. For numerical stability, the signed distance targets are truncated to a narrow band. The decoder parameters and the latent features of the involved (KNN) neural points are updated online by minimizing an SDF regression loss over these samples for a small fixed number of gradient steps per incoming scan, enabling efficient incremental submap accumulation without explicit voxel storage.



\subsection{Normal Vector Descriptor}
After getting the implicit submap representation, we extract the scene’s occupancy, the mesh's points, and their normal directions.

To extract this information, the process begins by discretizing the submap into a voxel grid with a predefined resolution $r$. At each voxel corner point within this grid, we query the Signed Distance Function (SDF) value according to the method described previously. Based on these SDF values, we then apply an enhanced Marching Cubes (MC) algorithm to reconstruct a triangular mesh. This algorithm generates vertices uniformly distributed on the reconstructed surface. Subsequently, we calculate the normal vector for each vertex, defining the local surface orientation. To further refine the mesh, we filter the vertices to remove isolated outliers, thereby retaining only significant vertices along with their normals. Finally, by analysing the SDF values at the corners of each voxel, we derive occupancy information: voxels exhibiting a sign change in their corner SDF values are classified as occupied. This occupancy classification is essential for distinguishing between occupied and free space in the environment.

Subsequently, inspired by the work of FEC~\cite{ref72,ref73}, we apply a fast clustering method for mesh vertices. Mesh vertices are grouped using radius-based Euclidean clustering accelerated by a KD-tree. Neighboring labels are progressively merged to obtain the dominant 3D clusters; implementation details are provided in the supplementary material.

    
    
    

Following the derivation of clusters $\mathbf{\mathcal{C} =\left \{ C_{1},...,C_{M} \right \}}$, points within each cluster are normalized. This normalization standardizes the scale of each cluster, thereby reducing sensitivity to noise and enhancing the robustness of subsequent geometric feature extraction. Specifically, all points in a cluster are isotropically scaled to be contained within a unit sphere of radius 1.

\begin{figure}[!t]
\centering
\begin{minipage}[b]{0.55\linewidth}
  \centering
  \includegraphics[width=\linewidth]{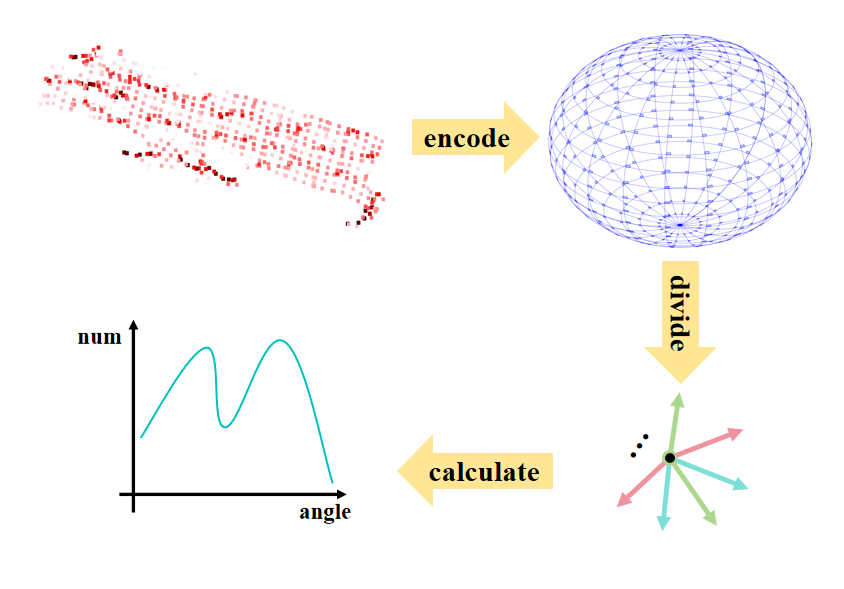}\\[-2pt]
  (a) 
\end{minipage}
\hfill   
\begin{minipage}[b]{0.4\linewidth}
  \centering
  \includegraphics[width=\linewidth]{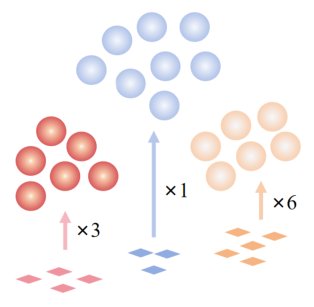}\\[-2pt]
  (b) 
\end{minipage}
\caption{(a) The figure illustrates the extraction of geometric descriptors for 3D clusters. The input is a 3D cluster with surface normals; color coding indicates normal deviations. First, these normals are mapped to discrete bins on a unit sphere, normalizing their distribution for systematic analysis. Then, the average angular deviation of the normals is calculated within each bin. Finally, these per-bin statistics are assembled into the geometric descriptors. (b) The figure illustrates the impact of keypoints on dominant 3-D clusters, where diamond markers denote macro-level keypoints and colored spheres represent the distinct constituent points at the micro-level. The cardinality of keypoints assigned to each dominant cluster subsequently governs the number of angular-distribution evaluations performed for that cluster.}
\label{fig_2}
\end{figure}

Following normalization, the next step involves partitioning the spherical domain into 72 sub-regions based on longitude and latitude. The sphere is first divided along the z-axis into northern and southern hemispheres. By latitude, each hemisphere is further divided into two bands, resulting in 4 latitude bands globally. Longitudinal division then subdivides each latitude band into 18 sectors, formed by meridians spaced 20 degrees apart. Fig.~\ref{fig_2}(a) illustrates this spatial subdivision.

Following the spherical partitioning, the mean normal vector is computed for each sub-region within a cluster. Next, the angular deviation between all unique pairs of these regional mean normal vectors within the same cluster is calculated. This process generates $\binom{n}{2}$ distinct angular measurements per cluster, where $n$ is the number of sub-regions (e.g., $n=72$). We then randomly select $n_{m}$ measurements. A histogram of these angular deviations is constructed per cluster using 10-degree bin intervals, resulting in $d=18$ dimensions. This process yields a $d$-dimensional geometric descriptor vector $\mathbf{F_{C}} \in \mathbb{R}^{d}$ for each cluster $C$.

The resulting angular-deviation histogram is used as the cluster-level normal descriptor, denoted as $\psi(C_m)$. The saliency-aware submap-level aggregation of these cluster descriptors is introduced in  Section \uppercase\expandafter{\romannumeral3}-E.


\subsection{BEV descriptor}
We project the occupancy evidence of the implicit submap onto the horizontal plane to obtain a BEV image $B$. The cluster-guided refinement of this projection is formalized in Section \uppercase\expandafter{\romannumeral3}-E. Here, we focus on extracting the BEV descriptor from $B$. The size of each pixel in the BEV matches the physical size of the occupancy grids. 
To capture orientation-aware structural information, we apply multi-scale and multi-orientation Log-Gabor~\cite{ref75} filtering to 
$B$. The filter responses are aggregated across scales, and the orientation with the maximum response at each pixel is recorded in a maximum index map (MIM)~\cite{ref76}.
FAST is then applied to the MIM to detect salient BEV keypoints. For each keypoint, a local orientation is estimated from the surrounding MIM region. The local patch is rotated according to this orientation and divided into regular subregions. Directional histograms extracted from these subregions are aggregated using VLAD~\cite{ref74}, yielding the global BEV descriptor $\mathbf{F^{B}}\! \in \!\mathbb{R}^{b}$. The detailed filter formulation, orientation assignment, patch normalization, and descriptor construction are provided in the supplementary material.

\subsection{Multi-Representation Geometric Hierarchy Fusion}
Given an implicit submap, our framework derives two complementary geometric representations: a macro-level BEV representation and a micro-level cluster-wise surface representation. Importantly, these two representations are not generated independently. Instead, they are coupled through a hierarchical interaction. The 3D clusters first constrain and refine the BEV generation process, while the resulting BEV keypoints subsequently guide the weighting and computational scheduling of micro-level cluster descriptors.

Let $\mathcal{O}$ denote the occupancy evidence extracted from the implicit submap, and let $\mathbf{\mathcal{C} =\left \{ C_{1},...,C_{M} \right \}}$ be the set of dominant 3D clusters obtained from the mesh vertices. For each cluster $\mathbf{C_m}$, we define its spatial support region in the occupancy grid as:
\begin{equation}
\label{eq121}
\mathcal{V}_m={v\in\mathcal{O}\mid d(\mathbf{x_v},\mathbf{C_m)}\leq \delta_c},
\end{equation}
where $\mathbf{x_v}$ is the 3D center of voxel $v$, $d(\cdot,\cdot)$ denotes the Euclidean distance from a voxel center to a cluster, and $\delta_c$ is a distance threshold controlling the cluster-supported projection region. The union of all cluster-supported voxels is then given by:
\begin{equation}
\label{eq122}
\mathcal{V}_{C}=\bigcup_{m=1}^{M}\mathcal{V}m.
\end{equation}
Only voxels belonging to $\mathcal{V}_{C}$ are retained for BEV projection. Therefore, the cluster-guided BEV image is generated as:
\begin{equation}
\label{eq123}
\mathbf{B(u)}=\eta\left(
\sum_{v\in\mathcal{V}_{C}}
\mathbf{1}[\pi(v)=u]\cdot o_v
\right),
\end{equation}
where $\mathbf{u}$ denotes a BEV pixel, $\pi(\cdot)$ is the projection from the 3D voxel grid to the BEV plane, $o_v$ is the occupancy value of voxel $\mathbf{v}$, and $\eta(\cdot)$ is a normalization function that maps the projected occupancy evidence to the image intensity range. This cluster-guided projection suppresses isolated noisy voxels, weakly structured ground-surface responses, and spurious observations that do not belong to dominant 3D structures. Consequently, the BEV image emphasizes stable structural layouts rather than uniformly projecting all occupancy evidence.

After obtaining the cluster-guided BEV image $\mathbf{B}$, we extract BEV keypoints from the maximum-index map generated by Log-Gabor filtering. Let $\mathcal{K}^{B}=\left \{k_1,\ldots,k_{N_B}\right \}$ denote the detected BEV keypoints. Each BEV keypoint is then associated with one dominant 3D cluster according to its projected spatial support:
\begin{equation}
\label{eq124}
\phi(k_i)=\arg\min_{m\in{1,\ldots,M}} d(\Pi^{-1}(k_i),\mathbf{C_m}),
\end{equation}
where $\Pi^{-1}(k_i)$ denotes the 3D support region corresponding to the BEV keypoint $k_i$. As shown in Fig.~\ref{fig_2}(b), the macro-level saliency of the m-th cluster is estimated by counting the number of BEV keypoints assigned to it:
\begin{equation}
\label{eq125}
s_m=\sum_{i=1}^{N_B}\mathbf{1}[\phi(k_i)=m].
\end{equation}

The normalized cluster saliency weight is computed as:
\begin{equation}
\label{eq125}
\alpha_m=\frac{s_m+\epsilon}{\sum_{j=1}^{M}(s_j+\epsilon)},
\end{equation}
where $\epsilon$ is a small constant used to avoid zero weights for clusters without assigned keypoints.

This formulation establishes a bidirectional hierarchy between the two representations. On the one hand, dominant 3D clusters determine which occupancy evidence is projected into the BEV image, thereby improving the structural reliability of the macro-level representation. On the other hand, the BEV keypoints extracted from this refined macro representation estimate the structural saliency of each 3D cluster, which is then used to guide micro-level geometric descriptor computation.

For each cluster $\mathbf{C_m}$, we compute a micro-level geometric descriptor $\mathbf{\psi(C_m)}$ from the angular distribution of regional mean normal vectors. The saliency-aware micro descriptor of the whole submap is obtained as:
\begin{equation}
\label{eq126}
\mathbf{F^{G}{hier}}=\operatorname{VLAD}\left \{ \alpha_m\mathbf{\psi(C_m)} \right \}.
\end{equation}
In this way, clusters that are more salient in the BEV layout contribute more strongly to the final micro-level descriptor, while less informative or noisy clusters are suppressed.

Finally, the macro BEV descriptor $\mathbf{F^B}$ and the saliency-aware micro descriptor $\mathbf{F^{G}{hier}}$ are fused as
\begin{equation}
\label{eq127}
\mathbf{F^{fuse}}=\operatorname{Norm}\left(
\left[
\lambda_B \mathbf{F^B},;
\lambda_G \mathbf{F^{G}{hier}}
\right]
\right),
\end{equation}
where $[\cdot,\cdot]$ denotes feature concatenation, $\lambda_B$ and $\lambda_G$ control the relative contributions of macro and micro-level descriptors, and $\operatorname{Norm}(\cdot)$ denotes $L_2$ normalization. The similarity between two submaps $S_q$ and $S_r$ is computed using cosine similarity:
\begin{equation}
\label{eq128}
\operatorname{Sim}(S_q,S_r)=
\frac{(\mathbf{F^{fuse}_q})^\top \mathbf{F^{fuse}_r}}
{|\mathbf{F^{fuse}_q}|_2|\mathbf{F^{fuse}_r}|_2}.
\end{equation}

The proposed fusion strategy differs from simple descriptor concatenation. It first uses dominant 3D clusters to generate a cleaner and more structurally meaningful BEV representation, and then uses the BEV-derived saliency to schedule and weight the cluster-level normal-vector descriptors. This closed macro–micro interaction enables the final descriptor to preserve both global spatial layout and local surface geometry while reducing the influence of noisy, dynamic, and weakly discriminative regions.
After retrieving a loop candidate, the intermediate BEV orientation statistics provide a coarse SE(2) initialization, followed by ICP refinement. Detailed formulations, evaluation protocols, and quantitative pose-estimation results are provided in the supplementary material.

\section{EXPERIMENTAL RESULTS AND ANALYSES}
In this section, Section \uppercase\expandafter{\romannumeral4}-A describes the dataset selected by us and some experimental settings, Section \uppercase\expandafter{\romannumeral4}-B introduces the evaluation metrics of the experiment, and Section \uppercase\expandafter{\romannumeral4}-C presents the experimental results. Section \uppercase\expandafter{\romannumeral4}-D and Section \uppercase\expandafter{\romannumeral4}-E show the ablation experiment and running time and storage efficiency.

\subsection{Dataset and Experimental Settings}
Below, we provide a detailed description of the dataset used in this study. Our dataset includes various types of LiDAR and different application scenarios. 
The configuration of the LiDAR sensors used in this dataset is shown in supplementary material.

\noindent1) \textit{KITTI Dataset}~\cite{ref77}: This dataset was collected at 10 Hz in an urban environment using a mechanical LiDAR (Velodyne HDL-64E) mounted on top of a vehicle. We selected six sequences (00, 02, 05, 06, 07, 08) containing loop closure points for our experiments.

\noindent2)  \textit{KITTI-360 Dataset}~\cite{ref78}: This dataset was also collected at 10 Hz in an urban scenario using a LiDAR (Velodyne HDL-64E). We selected four sequences (00, 04, 05, 06) containing loop closure points for the experiment. 

\noindent3)  \textit{NCLT Dataset}~\cite{ref79}: This dataset was collected using a 32-line LiDAR (Velodyne HDL-32E) mounted on a Segway robot. It was primarily collected on the University of Michigan campus, capturing variations across the four seasons, differing lighting conditions, and dynamic objects. Based on loop closure availability, we selected four sequences for the experiment: 2012-05-26 (NCLT01), 2012-08-20 (NCLT02), 2012-09-28 (NCLT03), and 2013-04-05 (NCLT04). Additionally, sequences 2012-03-17, 2012-02-04, and 2012-08-20 were used for cross-temporal location recognition experiments.

\noindent4)  \textit{MulRan Dataset}~\cite{ref80}: This dataset contains data collected from multiple scenarios in South Korea using an Ouster OS1-64 LiDAR. We selected sequences 01 and 02 from the DCC, KAIST, and Riverside scenes for the experiment; these were labeled MulRan01 to MulRan06, respectively (e.g., DCC01=MulRan01, DCC02=MulRan02, etc.). Additionally, sequences Sejong01 to Sejong03 were used to conduct cross-temporal location recognition experiments.

Our feature extraction method is manually applied to the submaps derived from the accumulated point clouds. To benchmark our approach, we selected six handcrafted feature extraction methods for comparison. These were evaluated on the selected datasets. The methods are Scan Context++~\cite{ref56}, M2DP~\cite{ref54}, NDT~\cite{ref81}, BoW3D~\cite{ref82}, Ring++~\cite{ref83}, and STD~\cite{ref18}. M2DP, NDT, BoW3D, and STD can be applied directly to the accumulated submaps. To ensure a fair comparison, we also applied the Scan Context++ and Ring++ methods to the submaps. For this purpose, we re-projected the accumulated submap points onto the middle frame of the submap. This step simulated a dense "scan" captured from that pose. BoW3D's default configuration targets the KITTI dataset; therefore, we conducted our experiments only on the KITTI and KITTI-360 datasets. All experiments were conducted on a laptop equipped with an Intel Core i7-10875H CPU at 5.10 GHz and 16GB RAM. We utilized an RTX 2060 GPU for acceleration during the construction of the neural implicit representations. The parameter values used in our method are listed in supplementary material.

\subsection{Performance Evaluation Metrics}
In this article, the conditions for establishing a loop closure are defined as follows. For a query submap, first compute the average position of all its frames to represent the submap's position. If the distance between this position and that of a previous submap is less than or equal to 20 meters, and their indices differ by more than 50, then these two submaps are considered positive (loop closure) pairs. The 20-meter distance threshold was chosen based on the urban environment, the typical range of LiDAR sensors, and the overall scale. Each method outputs a candidate frame with its similarity score and compares the score to a predefined decision threshold. If the score exceeds the threshold, classify the match as positive; if not, classify it as negative. For a match classified as positive, compute the actual geometric distance between the query frame and the candidate frame. If this distance is less than 20 meters, it is considered a true positive (TP); otherwise, it is classified as a false positive (FP). For a match classified as negative, if no ground truth loop closure exists for the query frame (according to the above criteria), it is considered a true negative (TN). If a ground truth loop closure exists for the query frame, it is a false negative (FN).

The following is how the evaluation metrics we adopt are calculated:

\noindent1) \textit{Precision–Recall Curve:} In the field of place recognition, precision is defined as the ratio of true positive results to the total number of identified matches. Recall, in contrast, represents the ratio of true positive results to the total number of actual positive instances. These concepts can be formally expressed as:
\begin{equation}
\label{eq14}
Precision=\frac{TP}{TP+FP} 
\end{equation}
\begin{equation}
\label{eq15}
Recall=\frac{TP}{TP+FN} 
\end{equation}
Here, TP denotes the number of true positive outcomes, FP represents the number of false positive results, and FN indicates the number of false negative instances. The precision-recall curve is constructed by adjusting decision thresholds. Each point on this curve corresponds to a specific threshold setting, illustrating the precision-recall trade-off and visually demonstrating how these metrics co-vary with threshold adjustments.

\noindent2) \textit{AUC:} The Area Under the Curve (AUC) measures the area under the Precision-Recall (PR) curve. It is a common metric for checking model performance at different thresholds. AUC is a single number between 0 and 1. A higher AUC indicates that the model is more effective at identifying positive cases across various thresholds.

\noindent3) \textit{Max F1 Score:} The F1 score, calculated using Eq~\ref{eq16}, represents the harmonic mean of precision and recall. This score provides a balanced measure by combining both metrics. The Maximum F1 score acts as a comprehensive performance indicator. It is often used to identify the optimal decision threshold. By balancing the competing demands of precision and recall, it effectively reconciles these priorities. The peak F1 score value reflects the algorithm's optimal performance. It demonstrates a balance between accurately identifying true positives and minimizing false positives. 
\begin{equation}
\label{eq16}
F_{1}=2\times \frac{Precision\times Recall}{Precision+  Recall}  
\end{equation}

\begin{figure*}[!t]
    \centering
    \includegraphics[scale=0.4]{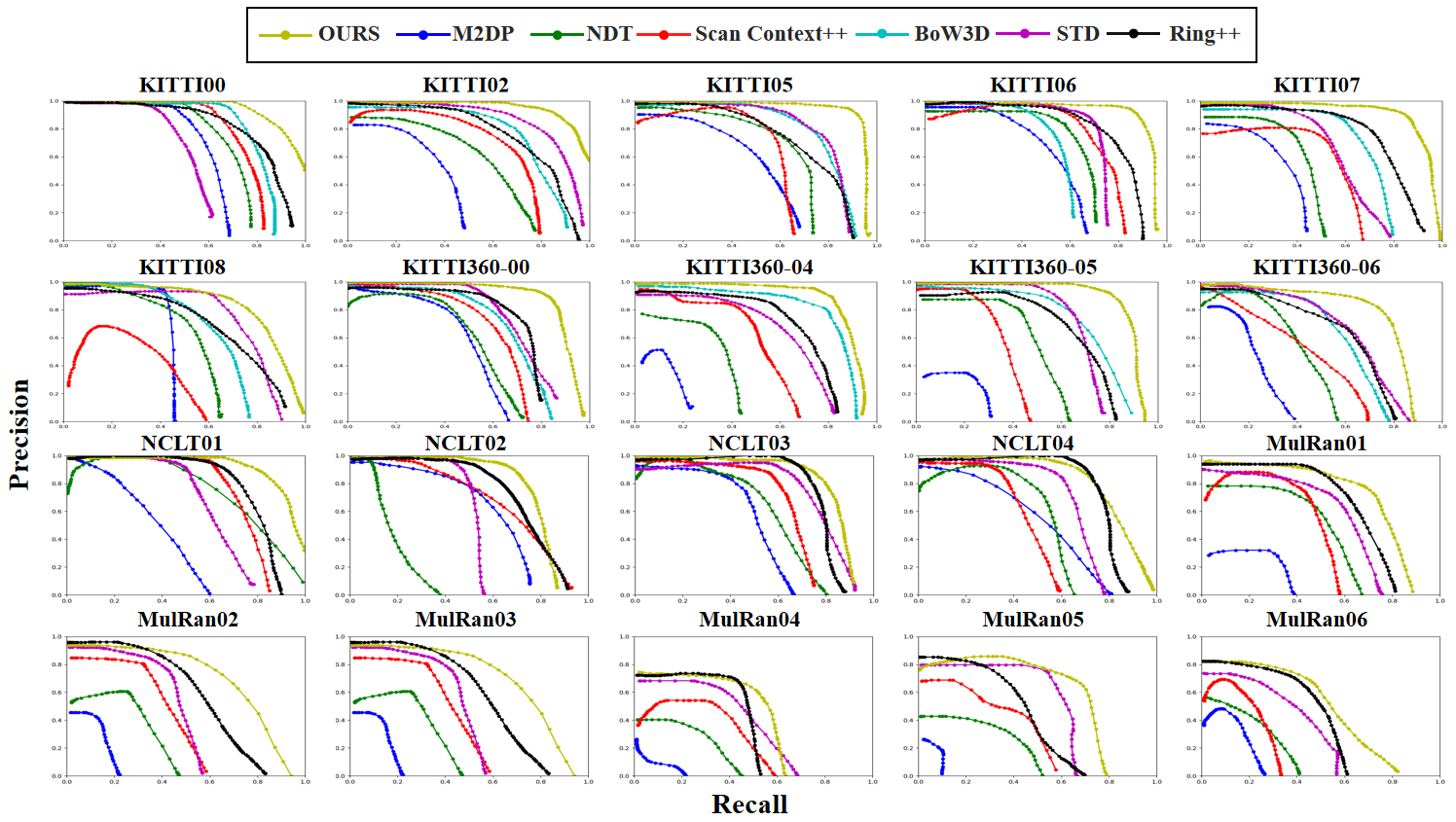}
    \caption{Evaluation of twenty short-term sequences reveals that each subfigure in Figure corresponds to the precision-recall performance of a specific method on these sequences. Our approach consistently outperforms all others across these sequences, demonstrating its robustness and adaptability. All image coordinates (x, y) are normalized to the unit interval [0, 1], and the subsequently figure PR curves adhere to the same convention.}
    \label{fig_4}
\end{figure*}

\begin{table*}[!t]
\caption{Comprehensive Performance Evaluation Across Multiple Datasets (AUC $\uparrow$ / Max F1 Score $\uparrow$ )}
\label{tab:table1}
\centering
\renewcommand{\arraystretch}{1.1}
\footnotesize
\begin{tabular}{@{}l *{10}{c} @{}}
\toprule
\multirow{2}{*}{\textbf{Method}} & 
\multicolumn{6}{c}{\textbf{KITTI}} & 
\multicolumn{4}{c}{\textbf{KITTI-360}} \\
\cmidrule(lr){2-7} \cmidrule(lr){8-11}
 & \rotatebox{45}{00} & \rotatebox{45}{02} & \rotatebox{45}{05} & \rotatebox{45}{06} & \rotatebox{45}{07} & \rotatebox{45}{08} &
 \rotatebox{45}{00} & \rotatebox{45}{04} & \rotatebox{45}{05} & \rotatebox{45}{06} \\
\midrule
M2DP & 
0.64/0.65 & 0.31/0.45 & 0.55/0.47 & 0.52/0.58 & 0.28/0.43 & 0.43/0.58 &
0.48/0.56 & 0.08/0.23 & 0.09/0.28 & 0.18/0.31 \\
\addlinespace[0.05cm]

NDT & 
0.69/0.70 & 0.52/0.59 & 0.61/0.67 & 0.60/0.68 & 0.38/0.52 & 0.52/0.60 &
0.51/0.57 & 0.26/0.44 & 0.44/0.56 & 0.39/0.48 \\
\addlinespace[0.05cm]

SC++ & 
0.72/0.75 & 0.63/0.67 & 0.55/0.65 & 0.69/0.73 & 0.47/0.60 & 0.27/0.44 &
0.61/0.64 & 0.58/0.47 & 0.35/0.45 & 0.42/0.48 \\
\addlinespace[0.05cm]

BoW3D & 
0.78/0.80 & 0.74/0.74 & 0.78/0.76 & 0.55/0.63 & 0.66/0.70 & 0.63/0.64 &
0.67/0.68 & 0.79/0.80 & 0.71/0.70 & 0.57/0.62 \\
\addlinespace[0.05cm]

STD & 
0.51/0.58 & 0.85/0.80 & 0.78/0.76 & 0.71/0.77 & 0.57/0.62 & 0.73/0.74 &
0.71/0.71 & 0.60/0.64 & 0.68/0.71 &  0.63/0.64 \\
\addlinespace[0.05cm]

Ring++ & 
0.80/0.76 & 0.76/0.71 & 0.71/0.68 & 0.80/0.76 & 0.75/0.75 & 0.67/0.65 &
0.70/0.73 & 0.67/0.69 & 0.62/0.66 & 0.59/0.63 \\
\addlinespace[0.05cm]

\textbf{Ours} & 
\textbf{0.93/0.86} & \textbf{0.95/0.88} & \textbf{0.93/0.91} & \textbf{0.91/0.89} & \textbf{0.92/0.87} & \textbf{0.90/0.85} &
\textbf{0.89/0.86} & \textbf{0.87/0.85} & \textbf{0.87/0.86} & \textbf{0.76/0.77} \\
\midrule
\midrule

\multirow{2}{*}{\textbf{Method}} & 
\multicolumn{6}{c}{\textbf{MulRan}} & 
\multicolumn{4}{c}{\textbf{NCLT}} \\
\cmidrule(lr){2-7} \cmidrule(lr){8-11}
& \rotatebox{45}{01} & \rotatebox{45}{02} & \rotatebox{45}{03} & \rotatebox{45}{04} & \rotatebox{45}{05} & \rotatebox{45}{06} &
 \rotatebox{45}{01} & \rotatebox{45}{02} & \rotatebox{45}{03} & \rotatebox{45}{04} \\
\midrule
M2DP & 
0.10/0.30 & 0.07/0.21 & 0.10/0.23 & 0.02/0.10 
 & 0.02/0.13 & 0.09/0.23 &
0.37/0.44 & 0.58/0.62 & 0.47/0.57 & 0.49/0.53 \\
\addlinespace[0.05cm]

NDT & 
0.41/0.55 &  0.21/0.38 & 0.17/0.34 & 0.13/0.30
& 0.18/0.35 & 0.15/0.30 &
0.75/0.69 & 0.18/0.26 & 0.57/0.61 & 0.51/0.61 \\
\addlinespace[0.05cm]

SC++ & 
0.41/0.55 & 0.36/0.48 & 0.34/0.48 & 0.24/0.42
& 0.29/0.44 & 0.17/0.32 &
0.74/0.74 & 0.67/0.64 & 0.63/0.69 & 0.44/0.52 \\
\addlinespace[0.05cm]

STD & 
0.55/0.62 & 0.43/0.55 & 0.36/0.50 & 0.34/0.48
& 0.49/0.63 & 0.32/0.44 &
0.61/0.65 & 0.51/0.62 & 0.74/0.75 & 0.64/0.70 \\
\addlinespace[0.05cm]

Ring++ & 
0.63/0.66 & 0.57/0.59 & 0.42/0.53 & 0.35/0.53 
& 0.38/0.49 & 0.40/0.52&
0.78/0.77 & 0.72/0.71 & 0.77/0.79 & 0.77/0.79 \\
\addlinespace[0.05cm]

\textbf{Ours} & 
\textbf{0.70/0.73} & \textbf{0.70/0.69} & \textbf{0.50/0.58} & \textbf{0.40/0.56}
 & \textbf{0.59/0.67} & \textbf{0.47/0.55} &
\textbf{0.92/0.85} & \textbf{0.77/0.78} & \textbf{0.83/0.81} & \textbf{0.82/0.78} \\
\bottomrule
\end{tabular}

\vspace{0.5em}
\begin{minipage}{\linewidth}
\footnotesize 
\textit{Note}: This table presents the Area Under the Curve (AUC) and maximum F1 scores for all evaluated methods across multiple datasets. The best result for each dataset is highlighted in bold. 
\end{minipage}
\end{table*}

\begin{figure*}[!t]
    \centering
    \includegraphics[scale=0.36]{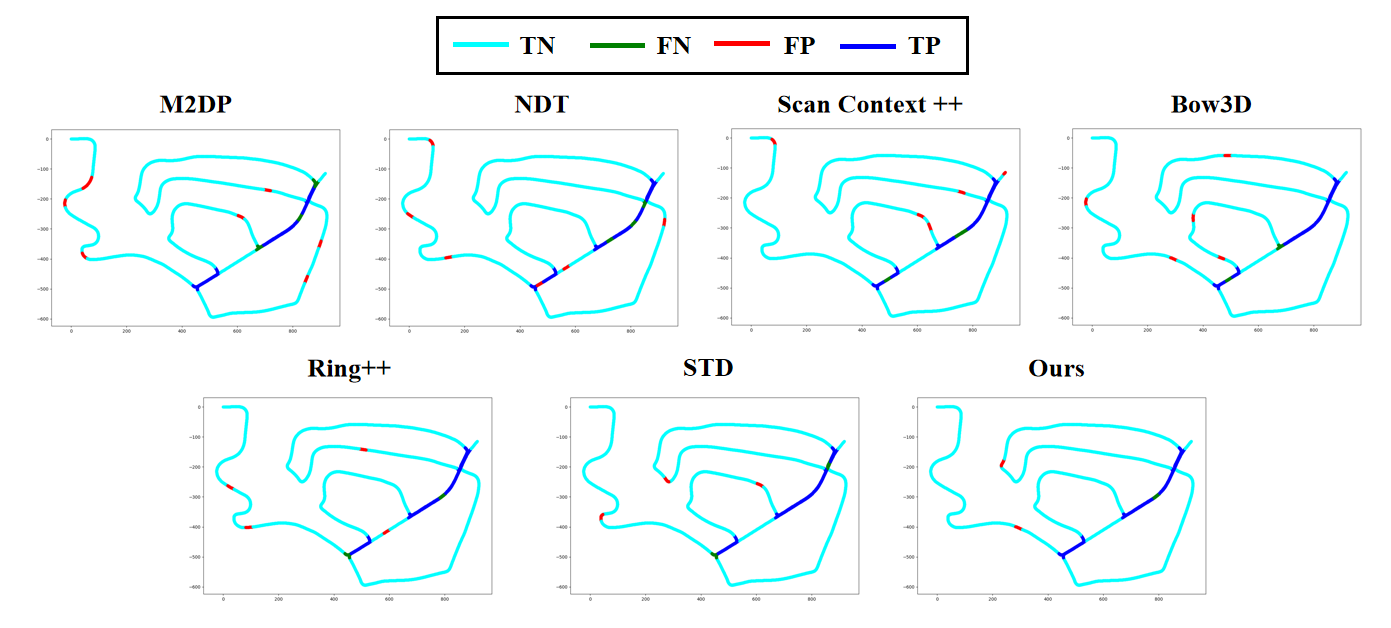}
    \caption{Loop retrieval result using each method’s max F1 score threshold on KITTI02. }
    \label{fig_5}
\end{figure*}

\begin{figure*}[!t]
    \centering
    \includegraphics[scale=0.7]{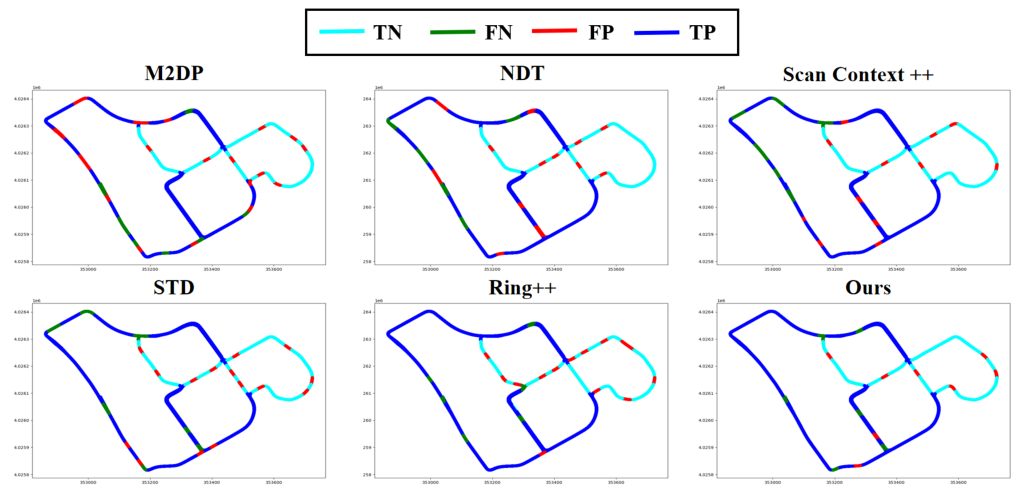}
    \caption{Loop retrieval result using each method’s max F1 score threshold on MulRan04.}
    \label{fig_6}
\end{figure*}

\subsection{Experiment Performance Comparisons}
To comprehensively test our method, we evaluated its performance in two major scenarios: short-term and long-term relocalization. Here, "short-term" refers to real-time detection of loop closure points. "Long-term" is the ability to recognize locations when revisiting the same area after more than one day. We further benchmark our approach against several learning-based counterparts. Our method ingests submaps accumulated from multi-frame point clouds. In contrast, most learning-based techniques operate on single-frame inputs. Therefore, we conduct the comparison under explicitly aligned data regimes to ensure fairness. For a rigorous comparison, we re-formulate the learning-based baselines (LCDnet\cite{ref84}, Bevplace++\cite{ref85}) so they ingest multi-frame inputs generated under the same conditions as our method. We adjust the temporal window and voxelization parameters so the tensors are shape-compatible with the pre-trained weights. This enables direct evaluation without retraining. Additionally, we conduct robustness experiments to evaluate the rotational and translational invariance of our method under viewpoint changes.

\noindent1) \textit{short-term:} The PR curves for our proposed method and existing methods are shown in Fig. \ref{fig_4}. We compared a total of 20 trajectories. Our method demonstrates superior performance compared to the existing methods. In the KITTI00 sequence, there are only small rotational and translational changes when passing through the same location, so most methods perform well. In contrast, scenarios with significant rotational and translational changes at the same location (such as KITTI05, KITTI08, KITTI360-04, and KITTI360-05) cause existing methods to generate inaccuracies. Our method retains robust performance in these cases. When there are numerous similar or repetitive scenes or when revisiting the same location multiple times, as in the six sequences in the MulRan dataset, all methods show some susceptibility to these challenging conditions. Nevertheless, our method still maintains good performance. Table \ref{tab:table1} presents the AUC values and peak F1 scores. Our method surpasses the compared approaches. Two sets of sequences for testing are presented in the figure to provide a detailed analysis of each method. All methods display their optimal performance, which is indicated by the maximum F1 score. Compared to other methods, ours achieves higher True Positives (TP) and lower False Positives (FP) and False Negatives (FN). To further highlight the distinction, we selected KITTI02 and MulRan04 tracks. We visualized their place recognition results, including TP, FP, FN, and TN, based on each method's optimal F1 score threshold. As shown in Figs. \ref{fig_5} and \ref{fig_6} show that our approach generates fewer false matches and detects loopback points more effectively. This delivers superior performance.

\begin{figure}[!t]
    \centering
    \includegraphics[scale=0.275]{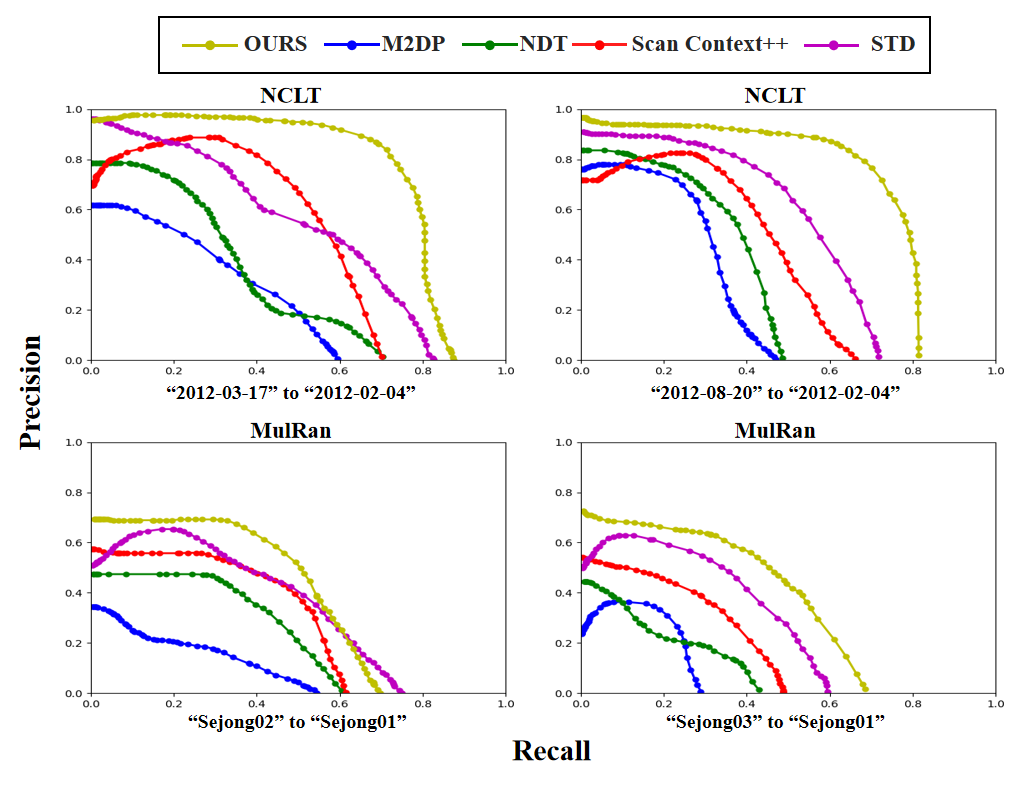}
    \caption{Precision–recall curve NCLT, MulRan on long-term.}
    \label{fig_7}
\end{figure}

\noindent2) \textit{long-term:} A robust place recognition system should perform reliably despite long-term environmental variations. To evaluate long-term place recognition performance, multiple intermittent datasets covering the same area were selected as the map sequence and query sequence, respectively. The results of our recall-precision experiments are shown in Fig.~\ref{fig_7}. AUC and F1 scores are presented in Table~\ref{tab2}. Our method achieves the best or competitive performance on most evaluated metrics. Although performance declines compared to short-term experiments on the same dataset, this decline is largely attributed to increased retrieval scope and significant environmental changes over time, such as seasonal differences, lighting shifts, or structural modifications. These factors make the long-term place recognition task more complex.

\begin{figure}[!t]
    \centering
    \includegraphics[scale=0.5]{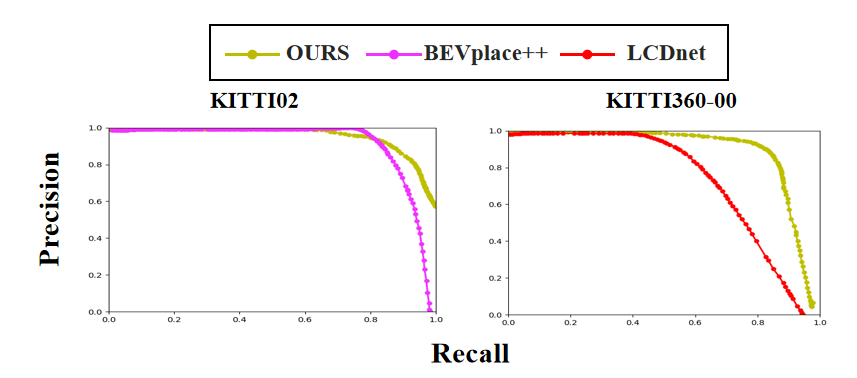}
    \caption{Precision–recall curve KITTI02, KITTI360-00 on learning-based method.}
    \label{fig_9}
\end{figure}

\noindent3) \textit{learning-based:} Although our framework leverages a learned implicit representation, the descriptor itself is constructed in a fully heuristic manner. This obviates the need for place-recognition labels and eliminates any training overhead. Nonetheless, to situate our approach within the state of the art, we benchmark it against leading learning-based baselines. The resulting PR curves are reported in Fig.~\ref{fig_9}. LCDNet releases weights exclusively for KITTI-360, and BEVPlace++ provides a reproducible training recipe on KITTI. We benchmark our method against BEVPlace++ on KITTI-02 and against LCDNet on KITTI-360-00, thereby adhering to the publicly available model domains. As shown, our method is competitive with the evaluated learning-based baselines under the adopted multi-frame submap setting. It performs better than LCDNet on KITTI-360-00 in this experiment, while BEVPlace++ remains stronger on KITTI-02, especially in the high-recall region. This clarifies the contexts in which our approach's performance is slightly lower, corroborating both its strong empirical efficacy and its untapped potential.
\noindent4) \textit{Robustness to view changes:} To further evaluate robustness under rotational perturbations, we randomly rotate all query and database point clouds around the z-axis within an angular range of ([0, 2$\pi$)), thereby simulating viewpoint variations commonly encountered in long-term autonomous driving scenarios without GPS. Experiments are conducted on KITTI06, KITTI07, NCLT03, and NCLT04, where submaps are consistently adopted as inputs to ensure fair comparison. As summarized in Table~\ref{tab:sequence_auc}, our method consistently achieves the best performance across all evaluation metrics. This result indicates that the proposed implicit 3D representation based on elastic neural points maintains stable and uniformly distributed structural encoding under arbitrary rotations. Moreover, the fused descriptors, which integrate bird’s-eye-view spatial layouts and cluster-level surface geometries, preserve complementary macro- and micro-level information despite rotational transformations. These findings verify the strong rotational invariance and structural robustness of the proposed framework in complex and dynamic environments.

As demonstrated, our method surpasses comparative approaches for four reasons. First, our implicit representation enhances BEV uniformity, producing more consistent spatial descriptors. Second, we integrate complementary geometric representations at both macro and micro scales, then fuse them geometrically, resulting in a richer and more hierarchical understanding of the environment. Third, our geometric descriptors disregard redundant ground-level data and focus on distinctive features, such as edges and corners, marking a clear methodological difference. Finally, the implicit representation suppresses dynamic elements, effectively reducing noise interference. Failure cases mainly arise from structural ambiguity between vegetation and buildings and from sparsified observations of distant structures; representative examples are provided in the supplementary material.

\begin{table}[!t]
\caption{Performance Comparison on Long-Term Sequences (AUC  $\uparrow$ / Max F1 Score  $\uparrow$ )}
\label{tab2}
\centering
\renewcommand{\arraystretch}{1.2}
\begin{tabular}{@{}l *{4}{c} @{}}
\toprule
\multirow{2}{*}{\textbf{Method}} & 
\multicolumn{2}{c}{\textbf{NCLT}} & 
\multicolumn{2}{c}{\textbf{MulRan}} \\
\cmidrule(lr){2-3} \cmidrule(lr){4-5}
 & \textbf{L1} & \textbf{L2} & \textbf{L1} & \textbf{L2} \\
\midrule
M2DP & 0.23/0.35 & 0.25/0.39 & 0.09/0.22 & 0.08/0.25 \\
\addlinespace[0.1cm]
NDT & 0.30/0.39 & 0.31/0.45 & 0.22/0.38 & 0.10/0.23 \\
\addlinespace[0.1cm]
SC++ & 0.49/0.57 & 0.39/0.49 & 0.29/0.45 & 0.18/0.34 \\
\addlinespace[0.1cm]
STD & 0.51/0.54 & 0.50/0.57 & 0.33/0.45 & 0.27/0.41 \\
\addlinespace[0.1cm]
Ring++ & 0.63/0.66 & 0.51/0.58 & 0.37/\textbf{0.52} & 0.31/0.48 \\
\addlinespace[0.1cm]
\textbf{Ours} & \textbf{0.75/0.77} & \textbf{0.70/0.73} & \textbf{0.38}/0.51 & \textbf{0.35/0.48} \\
\bottomrule
\end{tabular}

\vspace{0.5em}
\parbox{\columnwidth}{\footnotesize \textit{Note}: On the left is the AUC and on the right is the max F1 score. The best-performing method is bolded. "NCLT-L1" and "NCLT-L2" represent "2012-03-17" to "2012-02-04" and "2012-08-20" to "2012-02-04", respectively. MulRan-L1", and "MulRan-L2" represent "Sejong02" to "Sejong01 ", and" Sejong03" to "Sejong01 ", respectively.}
\end{table}

\begin{table*}[htbp]
\centering
\caption{Ablation Study Results: AUC $\uparrow$ and Max F1 $\uparrow$ Scores across Datasets}
\label{tab3}
\begin{tabular}{ccccccc}
\toprule
\multicolumn{3}{c}{\textbf{Components}} & \textbf{KITTI} & \textbf{KITTI-360} & \textbf{NCLT} & \textbf{MulRan} \\
\cmidrule(lr){1-3} \cmidrule(lr){4-4} \cmidrule(lr){5-5} \cmidrule(lr){6-6} \cmidrule(lr){7-7}
Implicit Rep. & BEV Image & Normal Vector & (AUC / F1) & (AUC / F1) & (AUC / F1) & (AUC / F1) \\
\midrule
\checkmark & \checkmark & \checkmark & \textbf{0.91 / 0.86} & \textbf{0.85 / 0.84} & \textbf{0.80 / 0.79} & \textbf{0.51 / 0.60} \\
 & \checkmark & \checkmark & 0.59 / 0.57 & 0.50 / 0.48 & 0.47 / 0.42 & 0.32 / 0.27 \\
\checkmark & \checkmark &  & 0.76 / 0.79 & 0.75 / 0.77 & 0.73 / 0.71 & 0.42 / 0.40 \\
\checkmark &  & \checkmark & 0.81 / 0.79 & 0.72 / 0.73 & 0.76 / 0.78 & 0.48 / 0.50 \\
\bottomrule
\end{tabular}
\vspace{0.2in}
\parbox{\linewidth}{\footnotesize \textit{Note}: The best performance for each dataset is highlighted in bold. Results are presented as AUC score / Max F1 score.}
\end{table*}







\begin{table}[!t]
\caption{Performance Comparison on Different Rotated Sequences AUC $\uparrow$/ Max F1 Score $\uparrow$}
\label{tab:sequence_auc}
\centering
\renewcommand{\arraystretch}{1.2}
\begin{tabular}{@{}l *{5}{c} @{}}
\toprule
\textbf{Method} & \textbf{KITTI06} & \textbf{KITTI07} & \textbf{NCLT03} & \textbf{NCLT04}  & \textbf{Mean} \\
\midrule
M2DP        & 0.39/0.48 & 0.16/0.30 & 0.26/0.38 & 0.24/0.35 &  0.26/0.38 \\
NDT     & 0.50/0.60 &  0.14/0.29 & 0.34/0.47 & 0.42/0.53 &  0.35/0.47 \\
SC++    & 0.38/0.52 & 0.26/0.40 & 0.41/0.53 & 0.24/0.35 & 0.32/0.45 \\
BoW3D   & 0.36/0.47 & 0.40/0.51 & \textbf{-} & \textbf{-} & 0.38/0.49 \\
STD   & 0.54/0.62 & 0.53/0.58 & 0.52/0.62 & 0.45/0.55 & 0.51/0.59  \\
Ring++    & 0.53/0.56 & 0.50/0.55 & 0.44/0.54 & 0.59/0.61 & 0.52/0.57  \\
\textbf{Ours}  & \textbf{0.87/0.84} & \textbf{0.78/0.74} & \textbf{0.59/0.63} & \textbf{0.68/0.70} & \textbf{0.73/0.73}  \\
\bottomrule
\end{tabular}

\vspace{0.5em}
\parbox{\columnwidth}{\footnotesize
\textit{Note}: The best performance in each column is highlighted in bold.}
\end{table}

\subsection{Ablation Study}
To analyze the contributions of individual method components to place recognition performance, we designed a series of ablation experiments. Specifically, we examined whether incorporating the implicit neural representation, as well as the BEV and normal vector modules, improves 3D place recognition accuracy.

To validate the contribution of the point-based implicit representation to our place recognition framework, we conducted two experimental configurations on our dataset. The first configuration retained all default parameters. The second configuration removed the implicit representation module. Without an implicit submap, we used a voxelization-based statistical method to generate occupancy grids for standard submaps. We calculated the point density within each voxel and set a density threshold that matched that of the implicit submaps. This ensured resolution consistency. For normal vector estimation, we used a principal component analysis (PCA)-based approach. PCA was computed on each point's local neighborhood. The eigenvector with the smallest eigenvalue was designated as the surface normal. Ablation tests were conducted across four datasets, encompassing both short-term and long-term scenarios. Quantitative results (AUC and Maximum F1-score) for all datasets are summarized in Table~\ref{tab3}.

Building on the implicit submap representation, we conducted two additional experiments. One used only BEV features, while the other used only normal vector features. Detailed quantitative comparisons are provided in Table~\ref{tab3}. As shown in the table, removing the implicit submap representation leads to significant performance degradation. This decline is due mainly to two sources: errors in surface normal estimation and inaccuracies in occupancy grid generation. Both introduce noise that compromises 3D place recognition robustness. The results also show that macro-level BEV information and local micro-scale information from the main 3D clusters work together to produce a robust descriptor. Additional ablations on dynamic filtering and submap construction distance are reported in the supplementary material.

\begin{table}
\begin{center}
\caption{TIME CONSUMPTION (MS) OF DIFFERENT METHODS}
\label{tab4}
\begin{tabular}{c||c c c c}
\hline
\textbf{Method} & \textbf{KITTI} & \textbf{KITTI360} & \textbf{NCLT} & \textbf{MulRan} \\
\hline
\textbf{M2DP}  & 73.2 & 74.1 & 80.2 & 76.2 \\
\hline
\textbf{NDT} & 465.2 & 451.2 & 387.2 & 392.1 \\
\hline
\textbf{SC++} & 62.1 & 59.4 & 57.2 & 63.1\\
\hline
\textbf{STD}& 38.2 & 33.2 & 36.9 & 40.1 \\
\hline
\textbf{Ring++}& 95.2 & 87.3 & 79.5 & 84.2 \\
\hline
\hline
\textbf{Ours} & 60.2 & 71.3 & 63.4 & 62.8 \\
\hline
\textbf{Ours(no implicit)} & 31.6 & 37.2& 33.9 & 35.2 \\
\hline
\textbf{Ours(only implicit)} & 28.6 & 34.2 & 29.5 & 27.6 \\
\hline
\end{tabular}
\end{center}
\end{table}

\begin{table}
\begin{center}
\caption{Memory consumption in MB}
\label{tab5}
\begin{tabular}{c||c | c }
\hline
\textbf{Sequence} & \textbf{Raw point cloud} & \textbf{Ours}  \\
\hline
KITTI02 & 9002.3 & \textbf{90.1(1.0\%)}\\
\hline
KITTI06  & 2052.2   &  \textbf{18.58(0.9\%)}   \\
\hline
KITTI360-00  &  21210.3   & \textbf{255.3(1.1\%)} \\
\hline
NCLT02  &  15366   & \textbf{108.2(0.7\%)} \\
\hline
\end{tabular}
\end{center}
\end{table}

\subsection{Computational Cost and Storage Efficiency}
We conducted comparative experiments on the runtime of our method to rigorously evaluate its efficiency in 3D place recognition. First, we performed a theoretical time complexity analysis for each major module of our framework, including implicit representation construction, voxel-based indexing, structural segmentation, descriptor generation, and retrieval. This analysis provides a clear understanding of the computational characteristics and scalability of each component. Subsequently, we empirically measured the average computation time of each module across all sequences. 
The detailed breakdown of the per-module average runtime is reported in the supplementary material.

To ensure a fair and statistically meaningful evaluation, we further analyzed the average time required to generate descriptors for each submap over the entire dataset. In particular, we separately measured the runtime of the three most critical components and compared them against representative state-of-the-art methods under the same experimental settings. The overall running time of our method was computed as the sum of descriptor generation time and retrieval time. The aggregated results are presented in Table~\ref{tab4}. As shown in Table~\ref{tab4}, our method achieves competitive runtime performance and is not inferior to most existing approaches, even though it leverages implicit representations. Notably, the implicit representation adopted in our framework is lightweight and computationally efficient, avoiding the heavy overhead typically associated with neural implicit models. Furthermore, we employ a Voxel Hashing technique to efficiently index key structural points, which significantly reduces redundant optimization iterations and accelerates spatial querying. The proposed large-scale 3D structural segmentation strategy is also carefully designed for efficiency, enabling rapid extraction of dominant geometric components without introducing additional computational burden.

We further demonstrate the storage efficiency advantage conferred by our implicit neural-point representation factor. Prior work often focuses solely on descriptor extraction while retaining raw point-cloud formats for downstream map fusion and maintenance. On several representative sequences, we quantify the compression ratio between our implicit encoding and the raw point clouds. Table~\ref{tab5} shows that our representation yields substantial storage savings. This advantage unlocks significant potential for subsequent map-merging and edge-cloud collaborative mapping tasks.

\section{CONCLUSION}
In this paper, we propose a novel 3D place recognition method inspired by implicit representations. Taking only occupancy grids, uniform triangular points and globally consistent normal vectors as inputs, we leverage submap-based implicit representations to process the data and generate BEV images as well as clustered 3D clusters. We derive BEV image descriptors via Log-Gabor filters, and compute 3D cluster descriptors based on angular differences between their normal vectors. Rather than simple concatenation, the proposed fusion mechanism constructs a hierarchical macro–micro representation by using macro-level BEV saliency to schedule and weight micro-level normal-statistics descriptors. Experiments verify that coupling implicit submaps with macro–micro geometric descriptor fusion can significantly improve the robustness and efficiency of 3D place recognition, confirming that implicit submap representations provide useful geometric evidence for this task. For future work, we will further explore the potential of implicit representations in place recognition, including developing task-tailored representation paradigms, building a large-scale standardized benchmark dataset for fair cross-environment comparisons, and designing an end-to-end place recognition framework fully based on implicit representations.

\vspace{-33pt}
\begin{IEEEbiography}[{\includegraphics[width=1in,height=1.25in,clip,keepaspectratio]{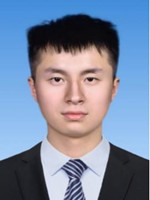}}]{Xiaohui Jiang}
received his B.S. and M.S. degrees from the College of Information Science and Technology, Beijing University of Chemical Technology (BUCT), Beijing, China, in 2023 and 2026, respectively. He is currently pursuing the Ph.D. degree at the College of Artificial Intelligence and Robotics, Hunan University, Changsha, China. His research interests include computer vision, SLAM and embodied intelligence.
\end{IEEEbiography}

\begin{IEEEbiography}[{\includegraphics[width=1in,height=1.25in,clip,keepaspectratio]{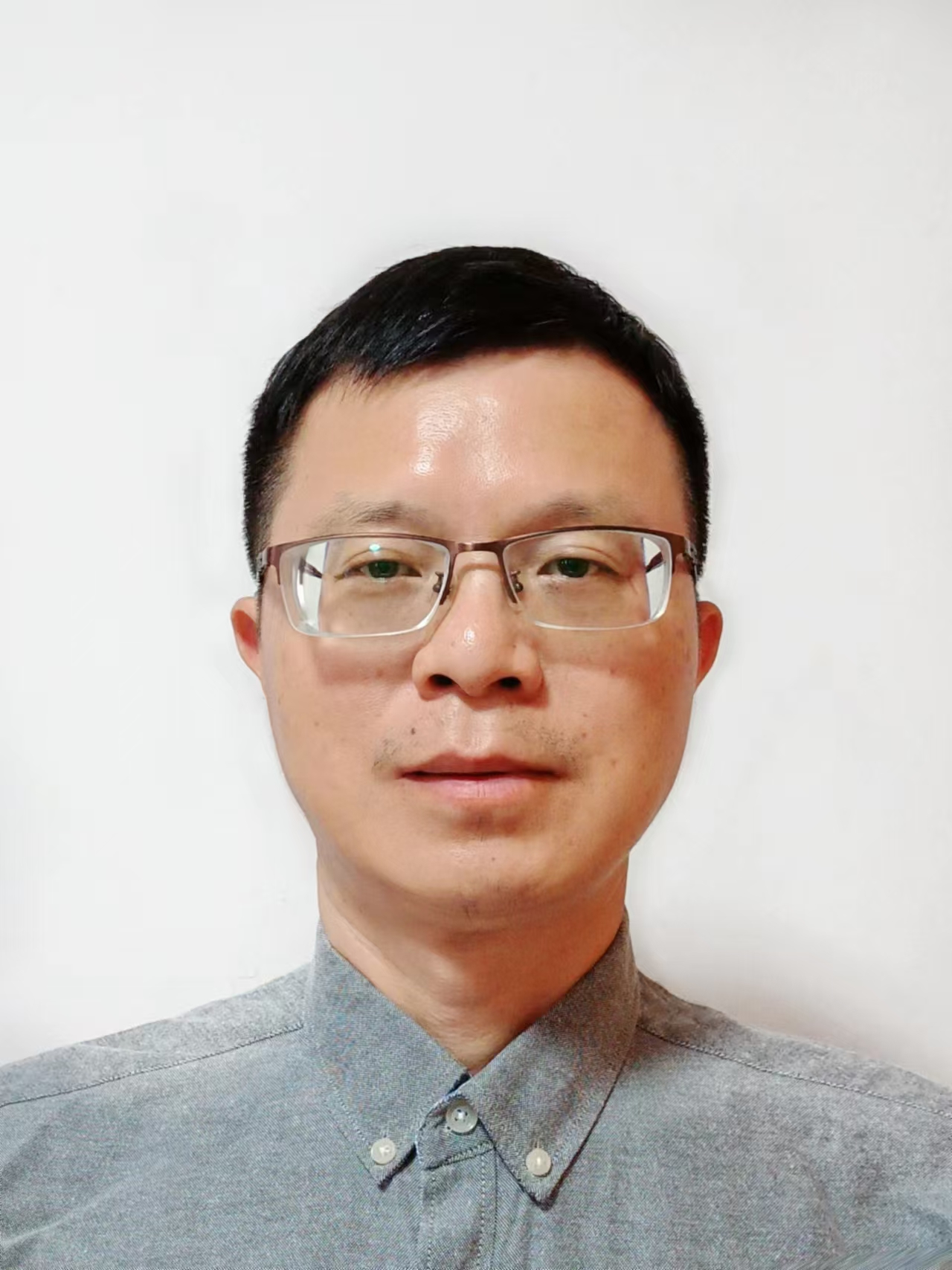}}]{Haijiang Zhu}
received the Ph.D. degree in pattern recognition and intelligent system from the National Laboratory of Pattern Recognition, Institute of Automation, Chinese Academy of Sciences, Beijing, China, in 2004. From 2006 to 2007, he was a visiting scholar at the Faculty of Engineering, Iwate University, Japan. Currently, he is professor and Ph.D. supervisor in the College of Information Science and Technology at Beijing University of Chemical Technology. His research interests include image processing and computer vision.
\end{IEEEbiography}

\begin{IEEEbiography}[{\includegraphics[width=1in,height=1.25in,clip,keepaspectratio]{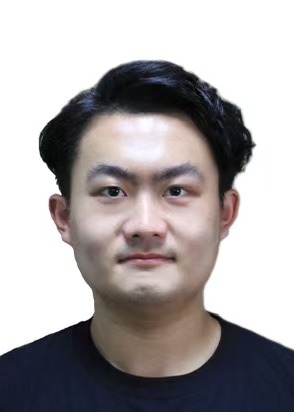}}]{Chade Li}
received the Ph.D. degree in pattern recognition and intelligent systems from the State Key Laboratory of Multimodal Artificial Intelligence Systems, Institute of Automation, Chinese Academy of Sciences, Beijing, in 2026. He is currently a Senior Engineer at Beijing Research Institute, Huawei Technologies Co, Ltd. His research interests include 3D vision and multi-modal perception and understanding.
\end{IEEEbiography}

\begin{IEEEbiography}[{\includegraphics[width=1in,height=1.25in,clip,keepaspectratio]{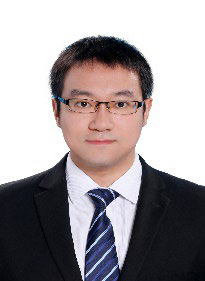}}]{Ning An}
received the Ph.D. degree in Control Theory and Control Engineering from the State
Key Laboratory of Management and Control for Complex Systems, Institute of Automation, Chinese
Academy of Sciences, in 2017, and the B.S. degree in Automation from China University of Mining and
Technology, in 2011. He is currently a Professor at Institute of Mining Artificial Intelligence, Chinese
Institute of Coal Science. His research interests include 3D perception for intelligent robot, 3D reconstruction of large-scale scenes, and multi-modal artificial intelligence systems.
\end{IEEEbiography}



\end{document}